\documentclass[11pt,a4paper]{article}
\usepackage[hyperref]{emnlp2018}
\usepackage{times}
\usepackage{latexsym}

\usepackage{url}

\usepackage{amsmath}
\usepackage{amssymb}
\usepackage{amsopn}
\usepackage{latexsym}
\usepackage{subcaption}
\usepackage{CJKutf8}
\usepackage{graphicx}
\usepackage{multirow}
\usepackage{color}
\usepackage{tabularx}
\usepackage{enumitem}
\usepackage{algorithm}  
\usepackage{algorithmic}  
\usepackage{caption}
\usepackage{capt-of}

\setlist[description]{font=\normalfont\itshape\space}

\usepackage{enumitem}

\aclfinalcopy 

\begin{document}
%
\title{Cseq2seq: Cyclic Sequence-to-Sequence Learning}
\author{Biao Zhang$^{1}$, Deyi Xiong$^{2}$ \textmd{and} Jinsong Su$^{1}$\\
	Xiamen University, Xiamen, China 361005$^{1}$ \\
	Soochow University, Suzhou, China 215006$^{2}$ \\
	{\tt zb@stu.xmu.edu.cn, dyxiong@suda.edu.cn, jssu@xmu.edu.cn} \\
}

\maketitle
\begin{abstract}

The vanilla sequence-to-sequence learning (seq2seq) reads and encodes a source sequence into a fixed-length vector only once, suffering from its insufficiency in modeling structural correspondence between the source and target sequence. Instead of handling this insufficiency with a linearly weighted attention mechanism, in this paper, we propose to use a recurrent neural network (RNN) as an alternative (Cseq2seq-I). During decoding, Cseq2seq-I cyclically feeds the previous decoding state back to the encoder as the initial state of the RNN, and reencodes source representations to produce context vectors. We surprisingly find that the introduced RNN succeeds in dynamically detecting translation-related source tokens according to the partial target sequence. Based on this finding, we further hypothesize that the partial target sequence can act as a feedback to improve the understanding of the source sequence. To test this hypothesis, we propose cyclic sequence-to-sequence learning (Cseq2seq-II) which differs from the seq2seq only in the reintroduction of previous decoding state into the same encoder. We further perform parameter sharing on Cseq2seq-II to reduce parameter redundancy and enhance regularization. In particular, we share the weights of the encoder and decoder, and two target-side word embeddings, making Cseq2seq-II equivalent to a single conditional RNN model, with 31\% parameters pruned but even better performance. Cseq2seq-II not only preserves the simplicity of seq2seq but also yields comparable and promising results on machine translation tasks. Experiments on Chinese-English and English-German translation show that Cseq2seq achieves significant and consistent improvements over seq2seq and is as competitive as the attention-based seq2seq model.

\end{abstract}

\section{Introduction}

Due to its unified model architecture and efficient end-to-end training schema, sequence-to-sequence learning (seq2seq) has attracted increasing attention from different communities and achieved very promising performance on various difficult problems, especially on  machine translation tasks in which neural machine translation (NMT) has achieved state-of-the-art results on multiple languages~\cite{DBLP:journals/corr/BahdanauCB14,DBLP:journals/corr/WuSCLNMKCGMKSJL16}. Current NMT models originate from the vanilla seq2seq, where an {\em encoder} encodes a source sequence into a fixed-length vector, from which a {\em decoder} generates its corresponding target sequence token by token~\cite{DBLP:journals/corr/SutskeverVL14,cho-EtAl:2014:EMNLP2014}.

\begin{figure*}[t]
\captionsetup[subfigure]{skip=12pt,belowskip=3pt,aboveskip=6pt}
\centering
  \begin{subfigure}{.3\textwidth}
  	\centering
  		\includegraphics[scale=0.48]{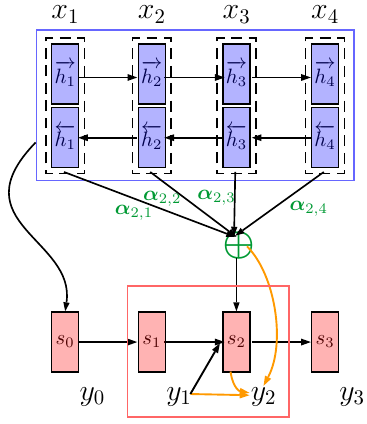}
  	\caption{\label{overall_vanilla}Attention-based Seq2seq}
  \end{subfigure}
  \begin{subfigure}{.3\textwidth}
  	\centering
  		\includegraphics[scale=0.48]{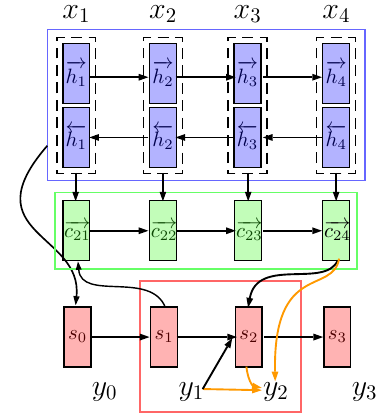}
  	\caption{\label{overall_rnmt}Cseq2seq-I}
  \end{subfigure}  
  \begin{subfigure}{.3\textwidth}
  	\centering
  		\includegraphics[scale=0.48]{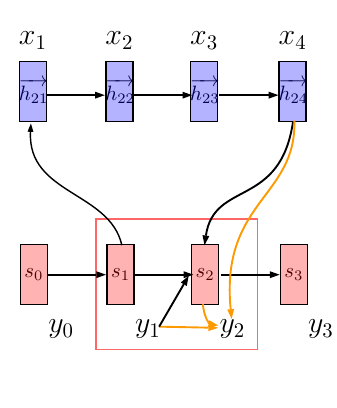}
  	\caption{\label{overall_seq2seq}Cseq2seq-II}
  \end{subfigure}  
\caption{\label{overall_v2} Illustration of the attention-based seq2seq model and Cseq2seq models. We use blue, green and red color to indicate encoder-related, attention-related and decoder-related representations respectively. The yellow lines show the flow of information for target token prediction.}
\end{figure*}

Unfortunately, this fixed-length vector is not sufficient to model the structural correspondence between the source and target sequence, resulting in unsatisfactory performance~\cite{DBLP:journals/corr/BahdanauCB14}. To address this issue,  Bahdanau et al.~\shortcite{DBLP:journals/corr/BahdanauCB14} introduce an attention mechanism (see Figure \ref{overall_vanilla}) which, as a bridge, makes the encoder and the decoder couple more tightly. In spite of its success, the underlying attention network is a weighted sum of encoded source representations, which is relatively simple and shallow.

In this paper, we propose to use a recurrent neural network (RNN) as an alternative to the attention network. The reasons are twofold: 1) RNN is able to handle variable-length inputs naturally~\cite{Hochreiter:1997:LSM:1246443.1246450} and 2) RNN, specifically the GRU model~\cite{journals/corr/ChungGCB14}, is capable of capturing complex and nonlinear dependencies inside its inputs. As shown in Figure \ref{overall_rnmt} which we refer to as Cseq2seq-I, we treat the target-side previous decoding state ${\mathbf s}_{j-1}$ as the initial state ${\mathbf c}_{j0}$ of the RNN, and feed all encoded source representations as its inputs. This enables the output of the RNN to change in accordance with the target-side translation history. Surprisingly, in practice we find that this RNN succeeds in dynamically detecting translation-related source tokens at each decoding step. 

Motivated by the success of Cseq2seq-I, we hypothesize that the partial target sequence can act as a feedback to improve the understanding of the source sequence, based on which we propose cyclic sequence-to-sequence learning (Cseq2seq-II, Figure \ref{overall_seq2seq}). Cseq2seq-II extends the vanilla seq2seq by merely reintroducing the previous decoding state into the encoder during decoding. In this way, the information from the decoder state is able to update or amend the understanding of the source sequence so that information flow between the encoder and the decoder becomes two-way. 
Thus, Cseq2seq-II can fully leverage what has been translated before and then guide the encoding procedure to focus more on those untranslated source tokens. This mixed understanding and translating behavior would benefit our model especially when the sequence is very long (e.g., a long sentence or a document).

Partially inspired by the recently proposed zero-shot translation~\cite{DBLP:journals/corr/JohnsonSLKWCTVW16,DBLP:journals/corr/HaNW16} where different languages use the same encoder and decoder, for Cseq2seq-II, we further share parameters between the encoder and decoder as well as between two sorts of target-side word embeddings\footnote{We treat weight matrix in softmax as another word embedding.}. This significantly prunes out 31\% parameters with even better performance, making our Cseq2seq-II equivalent to a single conditional RNN model. Cseq2seq-II is not only as simple as seq2seq in terms of architecture and training, but also efficient in inducing target-sequence-related source semantics. Even without the attention mechanism, Cseq2seq-II is still capable of capturing what should be translated in the next step as demonstrated in our experiments.

We conduct experiments on machine translation tasks over a variety of language pairs, including NIST Chinese-English and WMT English-German. The experimental results show that Cseq2seq achieves significant improvements over the vanilla seq2seq and obtains competitive results against the popular attention-based NMT.\footnote{Notice that although we verify our model on the machine translation task alone, our model is general enough to adapt to other sequence-to-sequence tasks.}

\section{Background: Attention-based Seq2seq}

This section briefly reviews the attention-based seq2seq which directly models the conditional translation probability $p(\mathbf{y}|\mathbf{x})$ of a target sequence $\mathbf{y}=\{y_1, \cdots, y_m\}$ given its corresponding source sequence $\mathbf{x}=\{x_1, \cdots, x_n\}$ as follows:
\begin{equation}
p(\mathbf{y}|\mathbf{x}) = \prod_{j=1}^{m} p(y_j|\mathbf{x}, \mathbf{y}_{<j})
\end{equation}
where $x_i,y_j$ are tokens in the source and target sequence respectively, and $\mathbf{y}_{<j} = \{y_1, \cdots, y_{j-1}\}$ is a partial translation. To predict the $j$-th target token $y_j$, attention-based seq2seq employs a non-linear neural network:
\begin{equation}\label{decoder_word}
p(y_j|\mathbf{x}, \mathbf{y}_{<j})=\text{softmax}\left(W_{p}g(E_{y_{j-1}}, \mathbf{s}_{j}, \mathbf{c}_{j})\right)
\end{equation}
where $W_p \in \mathbb{R}^{V_t \times d_w}$ is the weight matrix for target token prediction ($V_t$ is the size of target vocabulary), $E_{y_{j-1}} \in \mathbb{R}^{d_w}$ is the embedding of previous generated target token $y_{j-1}$, $\mathbf{s}_{j} \in \mathbb{R}^{d_h}$ is the $j$-th target-side hidden state (see Figure \ref{overall_vanilla}), $\mathbf{c}_j \in \mathbb{R}^{2d_h}$ is the translation-guided context vector and $g(\cdot)$ refers to a multilayer perceptron that projects the inputs into a $d_w$-dimensional semantic space. In this work, we treat $W_p$ as another type of target-side word embedding in addition to $E_y$.

Attention-based seq2seq uses a recurrent neural {\em decoder} to perform the above mentioned prediction: 
\begin{equation} \label{decoder_hidden}
\mathbf{s}_j = f_{dec}(\mathbf{s}_{j-1}, E_{y_{j-1}}, \mathbf{c}_j)
\end{equation}
The vanilla attention-based seq2seq directly uses a GRU to model $f_{dec}(\cdot)$. However, exploiting the information of $y_{j-1}$ before reading from the representation of source sequence should be beneficial, which has already supported in the latest {\em dl4mt} system\footnote{https://github.com/nyu-dl/dl4mt-tutorial/blob/master/\\session3/}. Therefore, we design $f_{dec}(\cdot)$ in two hierarchies:
\begin{align}
\mathbf{s}_j = \text{GRU}_2(\widetilde{\mathbf{s}}_j, {\mathbf{c}}_j), ~~\widetilde{\mathbf{s}}_j = \text{GRU}_1(\mathbf{s}_{j-1}, E_{y_{j-1}})
\end{align}
The decoder thus utilizes Eq. (\ref{decoder_hidden}) and (\ref{decoder_word}) alternatively to generate each token in the target sequence until reaching a predefined end token.

The context vector $\mathbf{c}_j$ is generated by the {\em attention mechanism}, a weighted sum of the encoded source representations:
\begin{equation}\label{attention_mechanism}
\begin{split}
\mathbf{c}_j = \sum\nolimits_i \alpha_{ji}\mathbf{h}_i \qquad \\
\text{with}~ \alpha_{ji} = \text{softmax}(e_{ji}),\qquad\qquad \\
 e_{ji} = v^T_a \tanh(W_a \widetilde{\mathbf{s}}_{j} + U_a \mathbf{h}_i)
\end{split}
\end{equation}
$\alpha_{ji}$ is the attention weight. Intuitively, it measures how well the output at position $j$ matches the inputs around position $i$. $\mathbf{h}_i$ is the encoded source representation at position $i$, which is calculated via a bidirectional recurrent {\em encoder}:
\begin{equation}\label{encoder}
\begin{split}
\mathbf{h}_i = [\overrightarrow{\mathbf{h}}_i, \overleftarrow{\mathbf{h}}_i ] \\
\overrightarrow{\mathbf{h}}_i = f_{enc}(\overrightarrow{\mathbf{h}}_{i-1}, E_{x_i}|\overrightarrow{\mathbf{h}}_0=\mathbf{0})\\
\overleftarrow{\mathbf{h}}_i = f_{enc}(\overleftarrow{\mathbf{h}}_{i+1}, E_{x_i}|\overleftarrow{\mathbf{h}}_n=\mathbf{0})
\end{split}
\end{equation}
where $E_{x_i} \in \mathbb{R}^{d_w}$ is the embedding for source token $x_i$, and $\overrightarrow{\mathbf{h}}_i, \overleftarrow{\mathbf{h}}_i \in \mathbb{R}^{d_h}$ are the hidden states generated in two directions. Following Bahdanau et al.~\shortcite{DBLP:journals/corr/BahdanauCB14}, we set the encoding function $f_{enc}(\cdot)$ to a GRU. 

\section{The Model}

In this section, we elaborate Cseq2seq-I and Cseq2seq-II. Both models use the same decoder as the attention-based seq2seq does. Therefore we focus on their differences in generating context vectors.

\subsection{Cseq2seq-I}

As we can see from Eq. (\ref{attention_mechanism}), the network for estimating the attention weights is relatively shallow, which may be inappropriate for dealing with complex nonlinear dependencies during decoding. We, instead, propose to use a RNN as the alternate to obtain context vectors. Particularly, we employ GRU as our recurrent unit
(see Figure \ref{overall_rnmt}):
\begin{equation}
\overrightarrow{\mathbf{c}}_{j,i} = f_{enc}(\overrightarrow{\mathbf{c}}_{j,i-1}, \mathbf{h}_{i}| \overrightarrow{\mathbf{c}}_{j,0} = \tanh(V \widetilde{\mathbf{s}}_{j} + b_0))
\end{equation}
Notice $f_{enc}=\text{GRU}$. Intuitively, the decoding state $\widetilde{\mathbf{s}}_{j}$ denotes the partial translation, which prompts the RNN to reread the source  sequence so as to put more emphasis on translation-relevant source tokens. This new information further guides the generation of the next target token. 

Comparing against the attention mechanism, the RNN uses deeper layers to encode two different sets of complex nonlinear dependencies: 1) the dependencies among encoded source representations $\{\mathbf{h}_i\}_{i=1}^n$, and 2) the dependencies between target translation history $\widetilde{\mathbf{s}}_{j}$ and source representations $\{\mathbf{h}_i\}_{i=1}^n$. All these dependencies are modeled with various nonlinear neural nets inside GRU, enabling our model to capture structural correspondence in-between the source and target sequences, which is especially beneficial for translation tasks.

{\bf Context Vector} We explore two approaches to compute the context vector from the hidden states $\mathcal{C}_j = (\overrightarrow{\mathbf{c}}_{j,1}, \overrightarrow{\mathbf{c}}_{j,2}, \ldots, \overrightarrow{\mathbf{c}}_{j,n})$, namely {\it mean-pooling} and {\it last-state}. The former uses the average of $\mathcal{C}_j$ as the context vector, which has proven effective in various classification tasks: $\mathbf{c}_j = \frac{1}{n} \sum_{i} \overrightarrow{\mathbf{c}}_{j,i}$. The latter uses the last hidden state, which is often assumed to have encoded all input information: $\mathbf{c}_j = \overrightarrow{\mathbf{c}}_{j,n}$. We conduct experiments to evaluate their effectiveness.

\subsection{Cseq2seq-II}

The success of Cseq2seq-I inspires us that the partial target sequence is able to prompt the RNN to detect translation-relevant source tokens so as to improve the understanding of the source sequence. We further propose Cseq2seq-II. Notice that in the vanilla seq2seq, the context vector $\mathbf{c}_j$ is always a fixed-length vector:
\begin{align}
\mathbf{c}_j = \overrightarrow{\mathbf{h}}_{n}, ~~
\overrightarrow{\mathbf{h}}_i = f_{enc} (\overrightarrow{\mathbf{h}}_{i-1}, E_{x_i}|\overrightarrow{\mathbf{h}}_0={\mathbf{0}}) \label{seq2seq_encoder}
\end{align}
Therefore, $\mathbf{c}_j$ has no relation with its generated partial translation. This is undesirable since compressing a source sequence into a single vector with a fixed dimensionality is insufficient to capture the varying translation context. Cseq2seq-II solves this problem by regarding the previous decoding hidden state $\widetilde{\mathbf{s}}_{j-1}$ as its translation context and using this state to initialize the same encoder to produce a translation-guided context vector:
\begin{equation}
\begin{split}
\mathbf{c}_j = \overrightarrow{\mathbf{h}}_{j,n} \\
\overrightarrow{\mathbf{h}}_{j,i} = f_{enc} (\overrightarrow{\mathbf{h}}_{j, i-1}, E_{x_i}|\overrightarrow{\mathbf{h}}_{j,0}=\phi(\widetilde{\mathbf{s}}_{j-1})) \label{cseq2seq_encoder}
\end{split}
\end{equation}
where $\phi(\cdot)$ denotes a one-layer feed-forward neural network. This cyclic encoder is initialized with the translation-aware decoder state. As a result, the encoder knows the current partial translation and what has been translated so far such that it can learn what should be translated next through controlling different gates to pay more attention to specific source tokens. 

We empirically find that the initialization of decoder state $\mathbf{s}_0$ in Cseq2seq-II is very important. Following Sutskever et al.~\shortcite{DBLP:journals/corr/SutskeverVL14}, we use the same encoder but in a reverse direction to build $\mathbf{s}_0$:\footnote{We did not choose the bidirectional RNN model here, because empirical experiments show no significant improvement in translation performance. However, the training time increases.}
\begin{align}
\mathbf{s}_0 = \phi(\overleftarrow{\mathbf{h}}_{0}), ~~ \overleftarrow{\mathbf{h}}_i = f_{enc} (\overleftarrow{\mathbf{h}}_{i+1}, E_{x_i}|\overleftarrow{\mathbf{h}}_n={\mathbf{0}}) \label{cseq2seq_initial}
\end{align}
In this way, the model introduces more short-term dependencies in the data that make the optimization problem much easier~\cite{DBLP:journals/corr/SutskeverVL14}. Notice that the $f_{enc}(\cdot)$ in Eq. (\ref{cseq2seq_encoder}) and Eq. (\ref{cseq2seq_initial}) are the same encoder using the same parameters but operate in different directions. 

 
\section{Parameter Sharing for Cseq2seq-II}

Deep neural models often suffer from over-parameterization, requiring large storage for a huge number of parameters~\cite{see2016compression}. In this paper, we explore two types of parameter sharing for Cseq2seq-II:
\begin{itemize}
\item
{\em Sharing GRU Parameters (SGRU)} As different languages can share the same encoder or decoder~\cite{DBLP:journals/corr/JohnsonSLKWCTVW16,DBLP:journals/corr/HaNW16}, we are curious about whether the encoder and decoder in Cseq2seq-II can share their GRU parameters. Concretely, we observe that the parameters in $f_{enc}$ (i.e. $E_{x}, \theta_{GRU}$) is a subset of that in $f_{dec}$ (i.e. $E_{y}, W_p, \theta_{GRU_1}, \theta_{GRU_2}$). Therefore, we share the parameters from $f_{enc}$ to $f_{dec}$, i.e. $\theta_{GRU} \to \theta_{GRU_1}$.
\item
{\em Sharing Word Embeddings (SWord)} We observe that there are two sorts of word embeddings in the target-side decoder, i.e. $W_p$ and $E_y$. Both matrices have the same shape and encodes the syntactic properties of target tokens. Following previous work~\cite{press-wolf:2017:EACLshort}, we use the same matrix for these two groups of word embeddings, i.e. $W_p \to E_y$.
\end{itemize}
This can bring the following two advantages: 1) sharing model parameters naturally reduces the parameter redundancy, and accordingly eases the requirement for physical memory; 2) forcing the parameters of different component to be the same encourages special model regularization, e.g. SWord enforces the NMT to not only capture syntactic and semantic properties of words but also utilize these properties to rank the candidate words for generating next word.
After sharing these parameters, there are only a source word embedding matrix $E_x$, a target word embedding matrix $E_y$, two feed-forward neural nets ($\phi(\cdot)$ in Eq. (\ref{cseq2seq_encoder}) and (\ref{cseq2seq_initial})) and a conditional GRU $\theta_{GRU_1}, \theta_{GRU_2}$. All together, our Cseq2seq-II is equivalent to use a single fully conditional RNN to perform the sequence-to-sequence learning tasks.
 
\section{Experiments}

We conducted a series of experiments on Chinese-English (Zh-En) and English-German (En-De) translation task to examine the effectiveness of the proposed model.

\subsection{Setup}

{\bf Datasets and Evaluation} For Zh-En, we collected 1.25M sentence pairs from LDC corpora\footnote{This corpora include LDC2002E18, LDC2003E07, LDC2003E14, Hansards portion of LDC2004T07, LDC2004T08 and LDC2005T06.} as our training data, which contains 27.9M Chinese words and 34.5M English words. We used the NIST 2005 (1082 sentences) dataset as our development set, and the NIST 2002 (878 sentences), 2003 (919 sentences), 2004 (1788 sentences), 2006 (1664 sentences), 2008 (1357 sentences) datasets as our test sets.


For En-De, we used the same WMT14 training data consisting of 4.5M sentence pairs with 116M English words and 110M German words\footnote{The preprocessed data can be found and downloaded from http://nlp.stanford.edu/projects/nmt/}~\cite{jean-EtAl:2015:ACL-IJCNLP,luong-pham-manning:2015:EMNLP,bojar-EtAl:2014:W14-33}. We used newstest2013 (3000 sentences) as our development set, and the newstest2014 (2737 sentences) and newstest2017 (3004 sentences) as the test sets. 

We measured the translation quality using case-insensitive (Zh-En) and case-sensitive (En-De) BLEU-4 metric~\cite{PapineniEtAl2002}, and performed paired bootstrap sampling~\cite{koehn04} for significance test. 

{\bf Baselines and Settings} We compared our model against three popular SMT and NMT systems: 
\begin{itemize}
\item {\it Moses}~\cite{Koehn:2007:MOS:1557769.1557821}: an open source conventional phrase-based SMT system. We equipped the system with a 4-gram language model trained on the target portion of training data using the SRILM toolkit with modified Kneser-Ney smoothing. Word alignment was generated by GIZA++ toolkit under the option ``{\em grow-diag-final-and}'', and the lexical reordering model was trained with the type ``{\em wbe-msd-bidirectional-fe-allff}''. 
\item {\it Seq2seq}~\cite{DBLP:journals/corr/SutskeverVL14}: the vanilla sequence-to-sequence learning model with only one layer RNN encoder and one layer RNN decoder.
\item {\it RNNSearch}~\cite{DBLP:journals/corr/BahdanauCB14}: the neural machine translation system enhanced with the attention mechanism. We implemented this model according to the publicly available {\em dl4mt}. Besides, we also provide the results of RNNSearch model that only has a single GRU layer in the decoder for reference. We refer to the latter one as {\it RNNSearch$^{-}$}.
\end{itemize}


{\bf Training Settings} We set the vocabulary size to 30K (Zh-En) and 40K (En-De). Byte pair encoding compression algorithm is used for En-De to handle the rich morphology issue~\cite{sennrich-haddow-birch:2016:P16-12}. Words that don't appear in the vocabulary were mapped to a special token {\em UNK}. We trained all NMT models with sentences of length up to 50 (Zh-En) and 80 (En-De) words in the training data, and set $d_w$ $=$ $620$, $d_h$ $=$ $1000$. We employed the Adadelta algorithm for optimization, with a batch size of $80$ and gradient norm as $5$. The initial learning rate was set to 1.0, and halved after each epoch. During decoding, we used the beam-search algorithm with a beam size of 10.


For all {\it Cseq2seq} models, we trained them from scratch, i.e. randomly initialized its parameters as in other NMT models. We implemented Cseq2seq based on the code of {\it dl4mt}. 

{\bf Complexity Analysis} Suppose the representation of a source sentence is $d$-dimensional with a length of $n$. At each decoding step, both RNNSearch and Cseq2seq have the same computational complexity of $\mathcal{O}(n \cdot d^2)$. The difference of them lies at the required number of sequential operation, where Cseq2seq needs $\mathcal{O}(n)$ operations compared with $\mathcal{O}(1)$ for RNNSearch. We further run all these models with Theano on GeForce GTX 1080 to check the practical effects. Within one hour, {\it RNNSearch} processes about 2800 batches, while our Cseq2seq-I and Cseq2seq-II process about 800 and 1000 batches respectively.

\begin{table*}[t]
\begin{center}
{\small
\begin{tabular}{l|c|c|lllllll}
\multicolumn{1}{l|}{\bf System} &
\multicolumn{1}{c|}{\bf \#Param} &
\multicolumn{1}{c|}{\bf MT05 } &
\multicolumn{1}{l}{\bf MT02 } &
\multicolumn{1}{l}{\bf MT03 } &
\multicolumn{1}{l}{\bf MT04 } &
\multicolumn{1}{l}{\bf MT06 } &
\multicolumn{1}{l}{\bf MT08 } &
\multicolumn{1}{l}{\bf ALL} \\
\hline
\hline
{\it Moses} & - & 31.70 & 33.61 & 32.63 & 34.36 & 31.00 & 23.96 & 31.03 \\

{\it Seq2seq} & 74.19M & 15.34 & 19.24 & 16.07 & 18.80 & 16.25 & 11.24 & 16.46 \\

{\it RNNSearch$^{-}$} & 84.35M & 31.24 & 36.29 & 34.24 & 37.37 & 32.63 & 25.34 & 33.34 \\

{\it RNNSearch} & 89.67M & 34.72 & 37.95 & 35.23 & 37.32 & 33.56 & 26.12 & 34.06 \\
\hline
\hline

{\it Cseq2seq-I ({\it MP})} & 90.05M & 33.79 & 36.19 & 34.45 & 37.38 & 32.91 & 24.94 & 33.38$^{\scriptsize \Uparrow}$ \\

{\it Cseq2seq-I ({\it LS})} & 90.05M & {\bf 36.79} & {\bf 39.18} & {\bf 37.18} & {\bf 39.86} & {\bf 35.61} & {\bf 27.41} & {\bf 36.00}$^{\scriptsize \Uparrow++}$ \\
\hline
\hline

{\it Cseq2seq-II} & 75.19M & 35.08 & 37.46 & 36.16 & 38.46 & 33.88 & 26.37 & 34.61$^{\scriptsize \Uparrow++}$ \\
{\it \qquad + SGRU} & 70.32M & 35.28 & 38.25 & 35.97 & 37.99 & 34.42 & 27.31 & 34.85$^{\scriptsize \Uparrow++}$ \\
{\it \qquad + SWord} & 56.59M & 35.62 & 38.05 & 35.89 & 38.92 & 34.06 & 26.68 & 34.89$^{\scriptsize \Uparrow++}$ \\
{\it \qquad + SGRU + SWord} & 51.72M & 35.33 & 38.18 & 36.45 & 38.64 & 34.38 & 26.23 & 34.89$^{\scriptsize \Uparrow++}$ \\
\hline
\end{tabular}
}
\end{center}
\caption{\label{all_test_performance} Case-insensitive BLEU scores on the Chinese-English translation tasks. {\bf \#Param} = the number of parameters in NMT model. {\it MP} and {\it LS} denote the {\it mean-pooling} and {\it last-state} respectively. {\bf ALL} = total BLEU score on test sets. We highlight the best results in bold for each test set. ``$\uparrow$/$\Uparrow$'': significantly better than {\it Moses} ($p$ $<$ $0.05$/$p$ $<$ $0.01$); ``$+$/$++$'': significantly better than {\it RNNSearch} ($p$ $<$ $0.05$/$p$ $<$ $0.01$).}
\end{table*}

\subsection{Results on Chinese-English Translation}

Table \ref{all_test_performance} summarizes the experiment results in terms of BLEU score. We find that the all NMT models achieve very promising results except the vanilla Seq2seq. This is mainly because that the vanilla Seq2seq model heavily relies on multi-layered deep encoders and decoders which are beyond our research scope. Our implemented RNNSearch model outperforms the Moses up to 3 points in total. Replacing the attention mechanism with GRU-based RNN model, Cseq2seq-I obtains another improvements of 1.94 points with the {\it last-state} strategy, and yields the best results across all test sets. Even without the attention mechanism, all Cseq2seq-II models outperform both RNNSearch and Moses system. All these improvements are significant. 

{\it Last-state vs. Mean-pooling for Cseq2seq-I.} It is interesting to observe that Cseq2seq-I with the {\it last-state} strategy yields significantly better performance than that with the strategy of {\it mean-pooling}, with a gain of 2.62 BLEU points in total. The reason behind is not completely clear, but can be revealed, to some extent, through the visualization of the GRU gates, where the mean-pooling operation may destroy the learned dependencies inside source sequences and in-between source and target sequences.
 This suggests that the {\it last-state} strategy for estimating context vectors is more suitable for Cseq2seq-I.

{\it Whether can the partial target sequence act as a feedback to improve the understanding of the source sequence?} Obviously, Cseq2seq-II obtains very competitive translation results against the RNNSearch, and significantly outperforms the Seq2seq by 18.15 BLEU points. Based on the fact that Cseq2seq-II only differs from Seq2seq in the feedback of the partial target sequence, we argue that the answer should be yes. As can be seen in the gates analysis, the partial target sequence guides the encoder to focus more on the translation-relevant source tokens. During decoding, this feedback help to improve the understanding of the source sentence toward adequate translation.

{\it Effects of Parameter Sharing for Cseq2seq-II.} To achieve similar results, RNNSearch uses 89.67M parameters, and Cseq2seq-I uses 90.05M parameters, while our Cseq2seq-II uses only 75.19M parameters (almost the same as that of Seq2seq 74.19M), with around 15M parameters pruned naturally. This illustrates the ability of Cseq2seq-II in using less parameters to model training data.

We then performed parameter sharing to verify whether there are parameters redundant in Cseq2seq-II. Firstly, we shared GRU parameters, which pruned 4.87M parameters. As shown in Table \ref{all_test_performance}, ``{\it Cseq2seq-II + SGRU}'' achieves 34.85 BLEU points in total, with a slight gain of 0.24 BLEU points. This suggests that different languages can indeed share their recurrent parameters, not only across the encoder or decoder~\cite{DBLP:journals/corr/JohnsonSLKWCTVW16,DBLP:journals/corr/HaNW16}, but also between the encoder and decoder. Secondly, we shared word embeddings, which pruned 18.6M parameters. Under this setting, ``{\it Cseq2seq-II + SWord}'' achieves 34.89 BLEU points, a gain of 0.28 BLEU points over Cseq2seq-II, indicating the two target-side word embeddings indeed share very similar syntactic properties and can be typed together. Finally, we combined the above two approaches, leading to a total of 23.47M parameters being pruned. With only 51.72M parameters, ``{\it Cseq2seq-II + SGRU + SWord}'' achieves 34.89 BLEU points, yielding 0.28 BLEU points improvement. In other words, we pruned 31\% parameters in Cseq2seq-II with no performance loss, which will benefit applications in memory-limited devices.


{\it Effects of Cseq2seq on Long Sequences.} Since RNN is able to model complex dependencies and is well known to encode long sequences, we carried out another group of experiments.

\begin{table}[t]
  	\centering
  	{\small
		\begin{tabular}{l||c}
		{\bf System} & {\bf AER} \\
		\hline
		\hline
		{\em Tu et al.~\shortcite{DBLP:journals/corr/TuLLLL16}} & 50.50 \\
		{\em RNNSearch} & 50.83 \\
		\hline
		{\em Cseq2seq-I (MP)} & 76.36 \\
		{\em Cseq2seq-I (LS)} & 57.43 \\
		\hline
		{\em Cseq2seq-II} & 70.49 \\
		{\em \qquad + SGRU} & 71.35 \\
		{\em \qquad + SWord} & 64.97 \\
		{\em \qquad + SGRU + SWord} & 67.88 \\
		\hline
		\end{tabular}
	}
	\caption{\label{align_quality} AER scores of word alignments deduced by different neural systems. The lower the score, the better the alignment quality.}
\end{table}

%

\begin{figure}[t]
  	\centering
  		\includegraphics[scale=0.35]{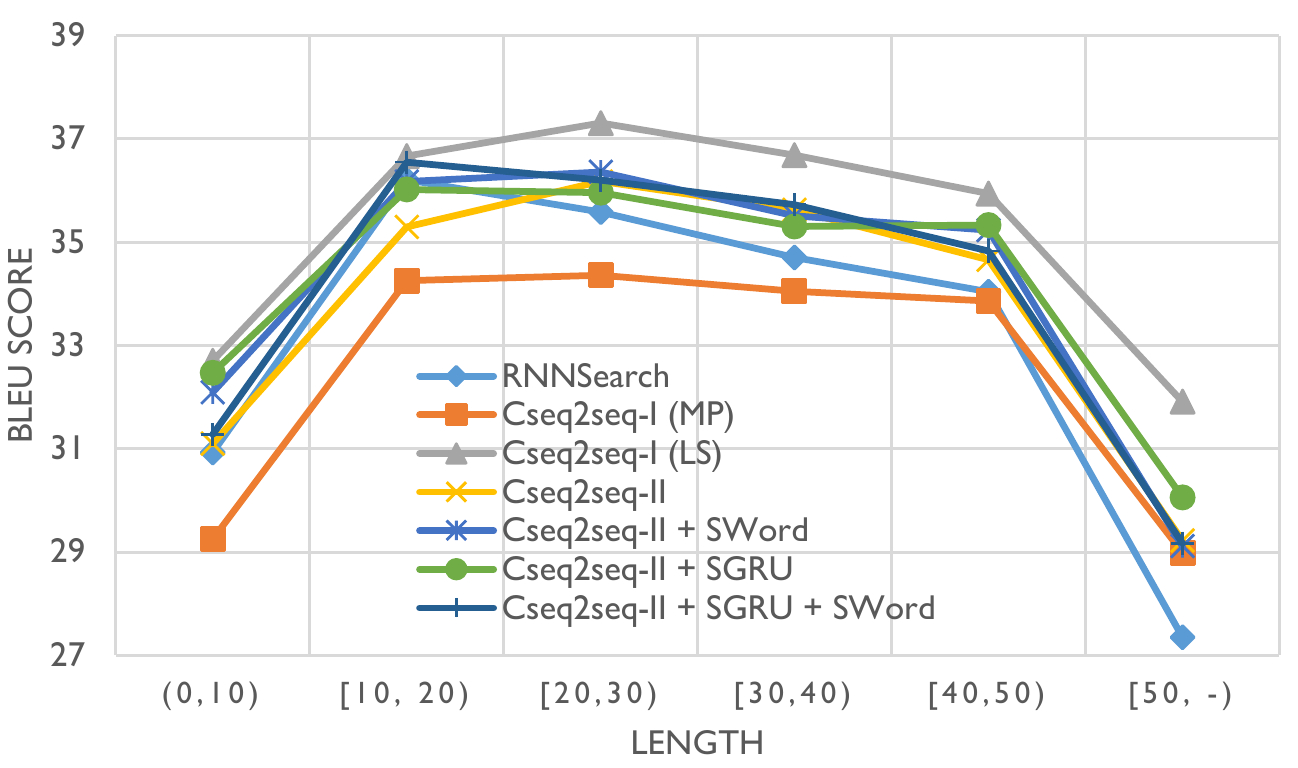}
  	\caption{\label{length_work} BLEU scores on different sentence groups according to their length.}
\end{figure}

Our second set of experiments testify whether Cseq2seq has better translation quality on long sentences in the original test sets. To this end, we divide our test sets into 6 disjoint groups according to the length of source sentences, each of which has 680, 1923, 1839, 1189, 597 and 378 sentences respectively. Figure \ref{length_work} illustrates the overall results. We find that as the length of source sentence exceeds a certain threshold (here over 50), the performance of NMT systems drops sharply, around 8 BLEU points. This indicates that long sentence translation is still a serious challenge for attention-based NMT systems, which resonates with findings of Tu et al.~\shortcite{DBLP:journals/corr/TuLLLL16} and Bentivogli et al.~\shortcite{2016arXiv160804631B}. However, compared with RNNSearch, Cseq2seq behaves more robust on translating long sentences, and all Cseq2seq models outperform RNNSearch on the longest sentences. Another notable observation is that Cseq2seq-I with the {\em mean-pooling} strategy generates worse results than the RNNSearch, but it outperforms the RNNSearch model on the longest sentence groups, a gain of 1.64 BLEU points. This further demonstrates that our model deals better with long sentences than the attention-based NMT.

\begin{table}[t]
\begin{center}
{ \small
\begin{tabular}{l|l|l}
\multicolumn{1}{l|}{\bf System } &
\multicolumn{1}{c|}{\bf newstest2014 } &
\multicolumn{1}{c}{\bf newstest2017 }\\
\hline
\hline
\it RNNSearch & 20.87 (20.78) & 22.33 (22.50) \\
\hline
\it Cseq2seq-I (MP) & 21.82 (21.58)$^{\scriptsize ++}$ & 22.90 (23.33)$^{\scriptsize ++}$   \\
\it Cseq2seq-I (LS) & {\bf 23.05} ({\bf 22.88})$^{\scriptsize ++}$ & {\bf 24.21} ({\bf 24.49})$^{\scriptsize ++}$ \\
\hline
\it Cseq2seq-II & 21.76 (21.65)$^{\scriptsize ++}$ & 23.10 (23.56)$^{\scriptsize ++}$ \\
\it \qquad + SGRU & 21.90 (21.77)$^{\scriptsize ++}$ & 23.26 (23.69)$^{\scriptsize ++}$ \\
\it \qquad + SWord & 22.54 (22.34)$^{\scriptsize ++}$ & 23.90 (24.31)$^{\scriptsize ++}$\\
\it \qquad + SGRU + SWord & 22.20 (22.02)$^{\scriptsize ++}$ & 23.54 (23.95)$^{\scriptsize ++}$ \\
\hline
\end{tabular}
}
\end{center}
\caption{\label{english_german_translation} Case-sensitive BLEU scores on the English-German translation task. We also provide the {\em detokenized} BLEU score in parentheses.}
\end{table}

\begin{figure*}[t]
  \centering
  \begin{subtable}{.24\textwidth}
  	\centering
  	\begin{tabular}{c}
  		\includegraphics[width=0.8\textwidth]{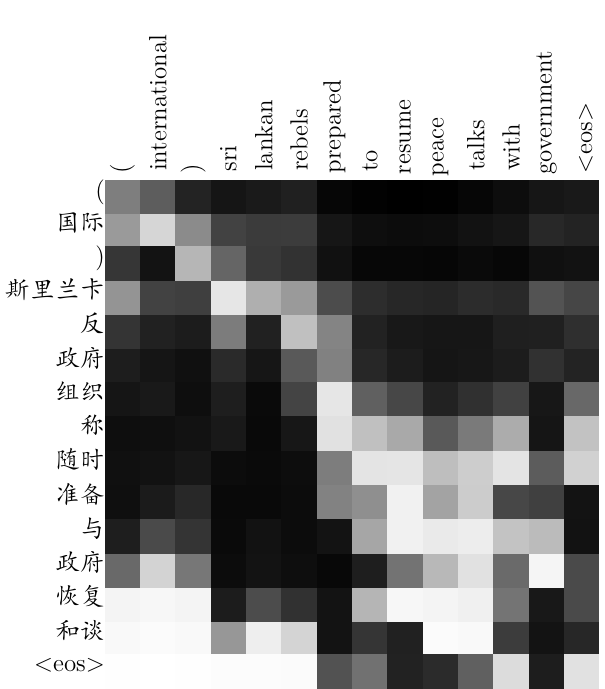}
  	\end{tabular}
  	\caption{Update gates (mean-pooling).}
  \end{subtable}
  \begin{subtable}{.24\textwidth}
  	\centering
  	\begin{tabular}{c}
  		\includegraphics[width=0.8\textwidth]{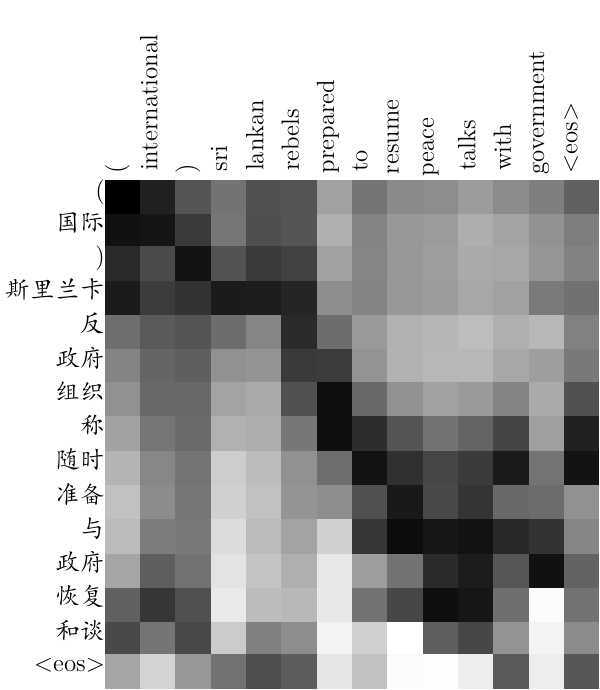}
  	\end{tabular}
  	\caption{Reset gates (mean-pooling).}
  \end{subtable}
  \begin{subtable}{.24\textwidth}
  	\centering
  	\begin{tabular}{c}
  		\includegraphics[width=0.8\textwidth]{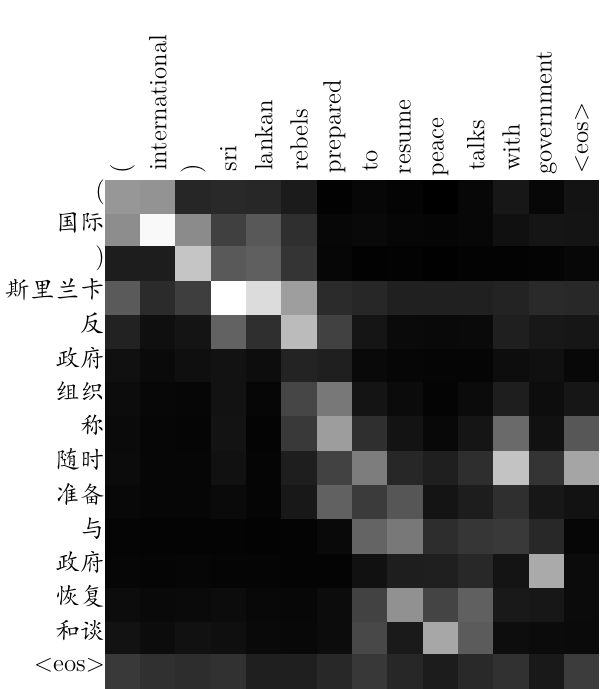}
  	\end{tabular}
  	\caption{Update gates (last-state).}
  \end{subtable}
  \begin{subtable}{.24\textwidth}
  	\centering
  	\begin{tabular}{c}
  		\includegraphics[width=0.8\textwidth]{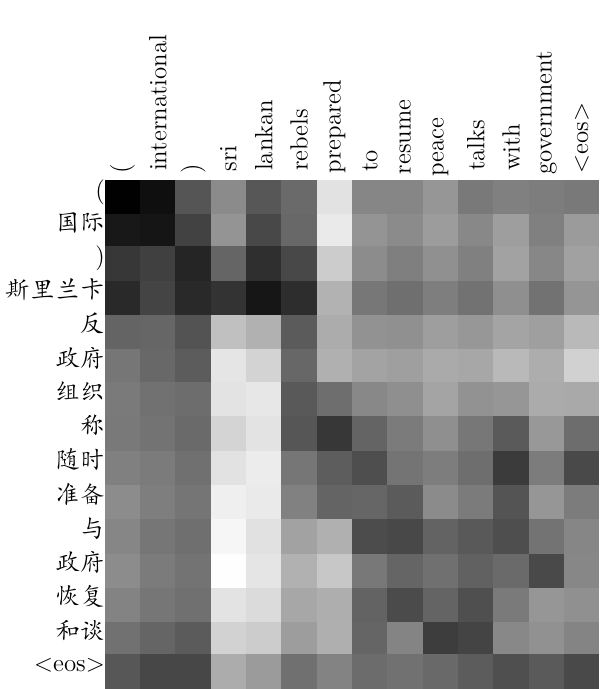}
  	\end{tabular}
  	\caption{Reset gates (last-state).}
  \end{subtable}
  \caption{\label{visual_work} Visualization of the gates in Cseq2seq-I. A pixel $(j, t)$ in the heatmap represents the reset/update gate metric $\widetilde{r}_{j,t}$, $\widetilde{z}_{j,t}$ of the $t$th source word and $j$th target word in grayscale (0: black, 1: white).}
\end{figure*}
\subsection{Results on English-German Translation}

Table \ref{english_german_translation} shows the translation results. We report both {\em tokenized} and {\em detokenized} BLEU scores, where the latter computes the evaluation metric using an in-built tokenizer and is consistent with WMT official BLEU results.
 In particular, the winning system in WMT14~\cite{BUCK14.1097.L14-1074}, a phrase-based system enhanced with a strong language model trained on huge monolingual text, yields a {\em detokenized} BLEU score of 20.70, which is slightly lower than our RNNSearch baseline (20.78).
After being equipped with our proposed recurrent mechanism, both Cseq2seq-I and Cseq2seq-II achieve significant improvements over the RNNSearch. Especially, Cseq2seq-II (LS) produces a {\em tokenized} BLEU score of 23.05, outperforming the RNNSearch with a substantial margin of 2.18 BLEU points. In addition, we have observed a similar phenomenon as in Chinese-English translation when model parameters are shared. This can be shown in the ``Cseq2seq-II + SWord'' which achieves a {\em tokenized} BLEU score of 22.54 with a gain of 0.78 BLEU points over the original Cseq2seq-II. And this can be also observed on WMT17 test set. All these indicate that 1) reducing the parameter redundancy benefits the translation performance and 2) our model can generalize to different language pairs with different-scale training corpus. 


\subsection{GRU Gate Analysis}

We are curious about the working mechanism inside Cseq2seq. To this end, we introduce a method to analyze the cyclic GRU RNN quantitatively.

The GRU consists of two gates: {\it reset} gate $\mathbf{r}$ and {\it update} gate $\mathbf{z}$. Intuitively, the reset gate controls the amount of information flowing from a previous state into a current state, while the update gate determines how much information should come from an input. That is, the reset gate can reflect the dependencies among source token representations, while the update gate can qualify how much each source token representation contributes to the target token prediction. Therefore, we define two metrics to analyze our cyclic RNN:
$
\widetilde{r}_{j,i} = \frac{1}{d_h} \sum\nolimits_{k=1}^{d_h} \mathbf{r}_{{j,i}_k}
,~~
\widetilde{z}_{j,i} = \frac{1}{d_h} \sum\nolimits_{k=1}^{d_h} \mathbf{z}_{{j,i}_k}
$, 
where $\widetilde{r}_{j,i}, \widetilde{z}_{j,i} \in (0, 1)$ due to the constraint of $\sigma(\cdot)$ in the gate function.

We visualize the heatmaps of $\widetilde{z}$ and $\widetilde{r}$ in Cseq2seq-I with {\it mean-pooling} and {\it last-state} for the same bilingual sentence pair in Figure \ref{visual_work}. We find that the update gates differ from the reset gates significantly, which suggests that these two gates learn different aspects of a sentence pair. In the update gate heatmap, it is very interesting to observe that the update gate can somewhat capture word alignments, with each pixel revealing a translation relation from the source to target language. In this perspective, the {\it last-state} strategy has a better capability of learning alignments than the {\it mean-pooling} strategy. Further quantitative analysis of the update gate on sufficiently many sentence pairs show that this gate is able to learn such translation attention. As for the reset gate, we find that it prefers to be opposite to the update gate. For example, the leading diagonals of the update gate heatmaps are white (value close to 1), while those of the reset gate heatmaps are black (value close 0). Additionally, the pixels for aligned word pairs seem to be turn points, which tend to have relatively small reset values. This is desirable, since previous hidden states on these points are blocked, while current hidden states are enabled to flow into final context vectors.

Secondly, we treated the update gate as word alignments, and evaluated its quality in terms of alignment error rate (AER) following Tu et al.~\cite{DBLP:journals/corr/TuLLLL16}.\footnote{Notice that we used the same dataset and evaluation script as Tu et al.~\cite{DBLP:journals/corr/TuLLLL16}. We refer the readers to \cite{DBLP:journals/corr/TuLLLL16} for more details.} We show the evaluation results in Table \ref{align_quality}. Consistent with the translation results, Cseq2seq-I with {\it last-state} strategy achieves the best alignment score. Notice that Cseq2seq does not aim at inducing high-quality word alignments. Instead, these AER results yielded by Cseq2seq demonstrate that the GRU gates inside Cseq2seq indeed learns what should be translated during translation.

\section{Related Work}

We focus on the advancements of sequence-to-sequence learning in NMT. Sutskever et al.~\shortcite{DBLP:journals/corr/SutskeverVL14} and Cho et al.~\shortcite{cho-EtAl:2014:EMNLP2014} propose the vanilla seq2seq model that only includes an encoder and a decoder. Soon, Bahdanau et al.~\shortcite{DBLP:journals/corr/BahdanauCB14} find that the fixed-length vector encoded in seq2seq is a bottleneck in improving its performance, and propose the attention-based seq2seq to automatically (soft-)search for parts of a source sequence that are relevant to predicting a target token. Luong et al.~\shortcite{luong-pham-manning:2015:EMNLP} further propose several different architectures, namely {\em global} and {\em local} model, for the attention mechanism so that the NMT can focus on more relevant parts on the source side during decoding. Different from these models, Cseq2seq-I uses a RNN as an alternative to the attention mechanism to model complex and highly nonlinear dependencies inside source sequences and in-between source and target sequences. Additionally, Cseq2seq-II significantly extends the seq2seq by dynamically rereading a source sentence at each decoding step. This cyclic encoding schema is very novel and efficient.

Cseq2seq is also related with memory networks~\cite{DBLP:journals/corr/WestonCB14,NIPS2015_5846} where the source sequence can be regarded as the read-only memory, and Cseq2seq uses a recurrent encoder to perform the reading operation (no writing operation). This differs significantly from the recent work of Wang et al.~\shortcite{wang-EtAl:2016:EMNLP20161} and Meng et al.~\shortcite{meng-EtAl:2016:COLING}. The former leverages an additional fixed-length memory to enhance the capacity of the decoder alone, while the latter treats the encoded source-side representations as an interactive memory. Both models perform both reading and writing operations, which make them more complex in training and design. From the view of memory networks, using recurrent models as the reading operation, to the best of our knowledge, has never been investigated before.

The idea of sharing the parameters of RNNs and word embeddings is partially motivated by the work of zero-shot translation~\cite{DBLP:journals/corr/JohnsonSLKWCTVW16,DBLP:journals/corr/HaNW16}. They find that different language pairs can share the same encoder or decoder without changing any structure of the NMT model, but the parameters of the encoder and decoder has to be separated. Different from their work, however, we share the parameters of the encoder and decoder, which to the best of our knowledge, is the first attempt. Our experimental results demonstrate its effectiveness, and also resonate with the findings of these zero-shot translations.

\section{Conclusion and Future Work}

In this paper, we have presented the cyclic sequence-to-sequence learning framework that leverages RNN to detect translation-relevant source tokens during decoding according to the partial target sequence. We observe that: 1) Cseq2seq can learn correspondences between the source and target language in an implicit way without any attention networks;
2) The encoder and decoder can share their RNN parameters without performance loss; 3) Cseq2seq is more effective than the traditional attention-based encoder-decoder in both Chinese-English and English-German machine translation.

In the future, we would like to test our approach on other sequence-to-sequence learning tasks, e.g. neural conversation system. Additionally, our model seems to be a non-proficient child translator who rereads the entire source sentence each time when a target word is generated. Rereading the whole source sentence for each target word is not necessary for a skilled translator. The source sentence will be reread only when cognitive difficulties occur during translation. We would like to model this process automatically using new neural models. Finally, we are also interested in applying our model to simultaneous translation.

\bibliography{emnlp2018}
\bibliographystyle{acl_natbib_nourl}

\end{document}